\documentclass[journal]{new-aiaa}
\usepackage[utf8]{inputenc}
\usepackage{textcomp}

\usepackage{subcaption}
\usepackage{graphicx}
\usepackage{amsmath}
\usepackage[version=4]{mhchem}
\usepackage{siunitx}
\usepackage{longtable,tabularx}
\usepackage{booktabs}

\title{Heat Transfer Prediction for Methane in Regenerative Cooling Channels with Neural Networks}

\author{G. Waxenegger-Wilfing \footnote{Research scientist, rocket engine department, Guenther.Waxenegger@dlr.de.}, K. Dresia \footnote{PhD student, rocket engine department, Kai.Dresia@dlr.de.}, J.C. Deeken \footnote{Group leader, rocket engine department, Jan.Deeken@dlr.de.} and M. Oschwald \footnote{Department head, rocket engine department, Michael.Oschwald@dlr.de.}}
\affil{DLR Institute of Space Propulsion, Hardthausen, Germany}

\begin{document}

\maketitle

\begin{abstract}
Methane is considered being a good choice as a propellant for future reusable launch systems. However, the heat transfer prediction for supercritical methane flowing in cooling channels of a regeneratively cooled combustion chamber is challenging. Because accurate heat transfer predictions are essential to design reliable and efficient cooling systems, heat transfer modeling is a fundamental issue to address. Advanced computational fluid dynamics (CFD) calculations achieve sufficient accuracy, but the associated computational cost prevents an efficient integration in optimization loops. Surrogate models based on artificial neural networks (ANNs) offer a great speed advantage. It is shown that an ANN, trained on data extracted from samples of CFD simulations, is able to predict the maximum wall temperature along straight rocket engine cooling channels using methane with convincing precision. The combination of the ANN model with simple relations for pressure drop and enthalpy rise results in a complete reduced order model, which can be used for numerically efficient design space exploration and optimization.
\end{abstract}

\section*{Nomenclature}


{\renewcommand\arraystretch{1.0}
\noindent\begin{longtable*}{@{}l @{\quad=\quad} l@{}}
$A$ & channel area $[\si{\milli \meter \squared}]$\\
$b$ & channel width $[\si{\milli \meter}]$\\
$d$ & wall thickness $[\si{\milli \meter}]$\\
$D_{h}$ & hydraulic diameter $[\si{\milli \meter}]$\\
$f$ & friction factor $[-]$\\
$G$ & mass flow density $[\si{\kilogram \per \second \per \meter \squared}]$\\
$h$ & channel height $[\si{\milli \meter}]$\\
$h$ & specific enthalpy $[\si{\kilo \joule \per \kilogram}]$\\
$ l$ & channel length $[\si{\milli \meter}]$\\
$\dot{m}$ & mass flow rate $[\si{\kilogram \per \second}]$\\
$p$ & pressure $[\si{\pascal}]$\\
$\rho$ & density $[\si{\kilogram \per \meter \cubed}]$\\
$\dot{Q}$ & heat flow rate $[\si{\watt}]$\\
$\dot{q}$ & heat flux $[\si{\watt \per \meter \squared}]$\\
$r$ & wall roughness $\si{[\micro \meter}]$\\
$Re$ & Reynolds number $[-]$\\
$T$ & temperature $[\si{\kelvin}]$\\
$v$ & flow velocity $[\si{\meter \per \second}]$\\
$y^+$ & dimensionless wall distance $[-]$\\
$z$ & flow length $[\si{\milli \meter}]$\\
\multicolumn{2}{@{}l}{Subscripts}\\
$b$ & bulk\\
$in$ & inlet\\
$out$ & outlet\\
$stat$ & static\\
$tot$ & total\\
$w$ & wall\\
\multicolumn{2}{@{}l}{Acronyms}\\
ANN & Artificial Neural Network\\
AR & Aspect Ratio \\
CFD & Computational Fluid Dynamics\\
CH$_{4}$ & Methane\\
MAE & Mean Absolute Error\\
ML & Machine Learning\\
ReLU & Rectified Linear Unit\\ 
RMS & Root Mean Square
\end{longtable*}}

\clearpage
\section{Introduction}
Although most liquid rocket engines that have flown until now used oxygen/hydrogen, oxygen/kerosene or a hypergolic propellant combination like nitrogen tetroxide/monomethylhydrazine \cite{Sutton2005, Caisso2009}, several countries started to develop engines that use methane as fuel and oxygen as oxydizer in the recent years. Oxygen/hydrogen offers the highest specific impulse, but the low density of hydrogen leads to large rocket stages. In addition, the low boiling temperature of hydrogen at \SI{20}{\kelvin} makes the handling very difficult and increases operating costs. Kerosene is much denser than hydrogen and easier to handle. Disadvantages are a lower specific impulse and that kerosene may coke and form deposits, which is problematic in terms of engine reuse. The main drawback of nitrogen tetroxide/monomethylhydrazine is its extreme toxicity. The propellant combination oxygen/methane has many favorable characteristics, e.g. methane is six times as dense as hydrogen, is easier to handle, has preferable coking temperature limits \cite{Liang1998} and low toxicity. Furthermore, oxygen/methane offers a slightly higher specific impulse than oxygen/kerosene \cite{Burkhardt2004}.

Despite the mentioned advantages, the prediction of heat transfer for methane flowing in the cooling channels of a regeneratively cooled combustion chamber has proven challenging \cite{Pizzarelli2015a}, but is needed for an efficient cooling system design. Regenerative cooling performance is especially important for engines, which are reusable or use an expander (bleed) power cycle, where the energy absorbed in the cooling channels drives the turbopumps \cite{Hahn2017, Leonardi2017}. The main difficulties for the heat transfer predictiont of methane is that it usually enters the cooling channels at supercritical pressure but subcritical temperature. It is then heated up in the cooling channels and most times crosses the Widom-line \cite{Banuti2015} close to the critical point. Strong changes in fluid properties at the Widom-line introduce various physical phenomena, e.g. heat transfer deterioration \cite{Urbano2012, Urbano2013}, which influence the heat transfer. In contrast to that, hydrogen usually enters the cooling system already in a gas-like state with pressures and temperatures far above the critical values \cite{Locke2008}. 

Several methods exist to study the regenerative cooling of liquid rocket engines. A simple approach is to use semi-empirical one-dimensional correlations to estimate the local heat transfer coefficient \cite{Dittus1985, Huzel1992}. By using an energy balance for each combustion chamber wall section, the wall temperatures can be estimated. Especially the maximum wall temperature, which occurs at the hot gas side, is a critical parameter, because it determines the fatigue life of the chamber \cite{Waxenegger2017}. The advantage of such simple relations is the negligible computation time. However, one-dimensional relations are not able to capture all relevant effects that occur in asymmetrically heated channels like thermal stratification \cite{Cook1984} or the influence of turbulence and wall roughness. Correction factors and quasi-two-dimensional models have been developed \cite{Pizzarelli2011}, but only full three-dimensional CFD calculations achieve convincing accuracy. Many papers have been published on CFD simulations for supercritical methane flowing in rocket engine cooling channels \cite{Ruan2012, Pizzarelli2012, Wang2013} and CFD results were compared with experimental data \cite{Haemisch2018, Haemisch2019}. The main disadvantage of CFD simulations is that they are not suitable for design optimization, design space exploration, and sensitivity analysis due to their large calculation effort.

By constructing surrogate models using samples of the computationally expensive calculation, one can alleviate this burden. However, it is crucial that the surrogate model mimics the behavior of the simulation model as closely as possible and generalizes well to unsampled locations, while being computationally cheap to evaluate. ANNs are known to be universal function approximators \cite{Hornik1989} and have been successfully applied as surrogate models in a number of domains \cite{Sudakov2019, Dresia2019}. Theses models have been applied to heat transfer prediction of supercritical fluids too \cite{Scalabrin2003a, Scalabrin2003, Chang2018}. The possibility to use ANNs with multiple hidden layers allows to generate surrogate models even for high-dimensional problems given a suitable number of samples. In this paper, for the first time, an ANN is trained with data extracted from samples of CFD simulations for heat transfer prediction of supercritical methane. The rest of the paper is organized as follows: Section \ref{sec:Artificial Neural Networks} describes the basics of machine learning (ML) and the theory of ANNs. A procedure to generate suitable training data by CFD calculations is presented in section \ref{sec:CFD based Data Generation}. Section \ref{sec:Artificial Neural Network for Wall Temperature Prediction} discusses the proposed ANN and reports the results. Section \ref{sec:Reduced Order Model for Cooling Channel Flow} shows how the ANN can be used as a building block of a complete reduced order model for cooling channel flows and section\ref{sec:Conclusion and Outlook} provides concluding remarks. A good deal of the material presented in the paper can also be found in the master thesis of one co-author supervised by the author \cite{Dresia2018}.

\section{Artificial Neural Networks}

\label{sec:Artificial Neural Networks}

ANNs are models that belong to the field of ML. To understand ANNs well, a basic understanding of the principles of ML is needed.  The following section briefly elaborates on the basics theory. A comprehensive presentation can be found in the book of Goodfellow et al. \cite{Goodfellow2017}.

\subsection{Machine Learning Basics}

The field of ML studies algorithms that use datasets to change parts of a mathematical model in order to solve a certain task, instead of using fixed pre-defined rules. The mathematical model is often a function, which maps input data to output data, and the algorithm has to learn the adjustable parameters of this function in such a way that the mapping has the desired properties.  In other words, ML is primarily concerned with the problem of finding and adjusting functions that usually have a large number of parameters. ML algorithms can be divided into supervised and unsupervised. In supervised learning, the training dataset contains both the inputs and the desired outputs, and the mathematical model can amongst other things be used for classification or regression. In a classification task, the model is asked to identify to which set of categories $k$ a specific input belongs. Assuming that each example of the input data is represented as a feature vector $\vec{x}\in\mathbb{R}^n$, the learning algorithm is asked to produce a suitable function $f: \mathbb{R}^n \to \{1,\dots,k\}$ with a discrete target output. A well-known example of a classification task is object recognition in images. In a regression task, e.g. with a single explanatory variable, the goal is to predict a numerical value given some input. To solve this task, the learning algorithm is asked to output a function $f: \mathbb{R}^n \to \mathbb{R}$. Unsupervised learning algorithms receive datasets without target outputs and learn useful properties of the structure of these datasets.

The central challenge in ML is that the model must perform well on new, previously unseen input data. The capability to perform well on those inputs is called generalization. Generalization is also central to understand the relationship between mathematical optimization and ML. While optimization algorithms can be used to minimize some error measure on the training set, ML tries to reduce the generalization error, also called the test error. During training, one must prevent two central issues. Underfitting occurs when the model is not able to obtain a sufficiently low error on the training data. Overfitting occurs when the gap between the training error and test error is too large; thus, the model is not able to generalize. The ability of a model to fit a wide variety of functions is called the model's capacity. Models with low capacity may have problems to fit the training data. Models with high capacity can solve complex tasks, but when their capacity is higher than needed, they may overfit by memorizing properties of the training data that do not work well on the previously unseen test data. ML achieves good results when the capacity of the model is appropriate for the true complexity of the relevant task and the amount of training data. However, for practical applications, it is nearly impossible to guess the model with an appropriate capacity. Furthermore, models with higher capacity in combination with proper methods to prevent overfitting often work better than less complex models. Modifications of a learning algorithm that are intended to reduce its generalization error, possibly by an increase in train error, are known as regularization. Instead of reducing the capacity of the model, one can, for example, change the learning algorithm to express the preference of one function over another.

Most ML models and algorithms have hyperparameters that are not adapted by the learning algorithm, but can be used to control the outcome, for example, by changing the capacity of a model. Optimal values of hyperparameters and estimates for the generalization error are found by splitting the available data into three disjoint subsets. The training set is used to adapt the trainable parameters by the learning algorithm. The second dataset, the validation set, exists to estimate the generalization error during or after training, allowing for hyperparameter tuning with the goal to find a good balance between performance and avoidance of overfitting. However, the estimate of the generalization error of the final model will be biased because the validation data was used to select the model. Thus, a third dataset, the test set, is used to estimate the real generalization error.

\subsection{Theory of Artificial Neural Networks}

One successful family of models used for ML is that of ANNs. ANNs are inspired by the functionality of biological brains, which are made of a huge number of biological neurons that work together to control the behavior of animals and humans. A collection of connected units, called artificial neurons, form the basis of an ANN. Furthermore, artificial neurons loosely model biological neurons and are usually represented by nonlinear functions acting on the weighted sum of its input signals. Let $\vec{in}=(in_{1},in_{2},\dots,in_{n})$  denote an input vector, $\vec{w}=(w_{1},w_{2},\dots,w_{n})$ a weight vector, where $n$ is the input dimension, $b$ a bias term and $\phi$ an activation function, then the output $out$ of a single artificial neuron can be written as

\begin{equation}
out=\phi\left(\sum_{j=1}^{n}w_{j}in_{j}+b\right).
\end{equation}
\label{eq:neuron_operation}

The bias term $b$ can be used to shift the activation function $\phi$.  A rectified linear unit (ReLU), where $\phi(x)=max\{0,x\}$, is the most common activation function in modern ANNs. Trainable parameters are usually the weights and biases of the neurons. Mostly, the connectivity architecture of such ANNs is layered with an input layer, multiple hidden layers and an output layer. ANNs are called feedforward networks when no feedback connections are present. One can prove that a feedforward network with a single hidden layer can approximate any reasonable function if the hidden layer has enough neurons. Nevertheless, using multiple hidden layers adds exponentially more expressive power. Amongst other things, each layer can be used to extract increasingly abstract features and hence more suitable representations of the input data. An ANN with more than one hidden layer is called a deep ANN. Such deep ANN can discover a suitable hierarchy of representations during training and as a result learn and also generalize better.  During training or learning, the algorithm requires a measure for the quality of its prediction to adjust the parameters of the model. In regression problems, a typical choice for the cost function is the mean squared error between predicted values and ground truth:

\begin{equation}
J(\vec{\theta})=\frac{1}{2m}\sum_{\vec{x}}\left(y(\vec{x})-f(\vec{x},\vec{\theta})\right)^2,
\end{equation}
\label{eq:cost}

where $\vec{x}$ and $y(\vec{x})$ are the input vector and the ground truth respectively of a training data point, $m$ is the total number of training data points and $f(\vec{x},\vec{\theta})$ is the predicted output of the model according to its parameters $\vec{\theta}$. For ANNs the model parameters $\vec{\theta}$ are given by all weights and biases associated to the neurons. Training correponds to finding optimal parameters $\vec{\theta}$ such that $J(\vec{\theta})$ is minimal. Often one adds an extra term for regularization:

\begin{equation}
\tilde{J}(\vec{\theta})=J(\vec{\theta})+\alpha \Omega(\vec{\theta})\quad\text{with}\quad\Omega(\vec{\theta})=\frac{1}{2}||\vec{w}||_{2}^{2}=\frac{1}{2}(w_{1}^2+w_{2}^{2}+\dots+w_{i}^{2}),
\end{equation}
\label{eq:cost_regularized}

where $\vec{w}$ denote the weights of the ANN, $i$ is the total number of trainable weights and $\alpha$ is an additional hyperparameter, which controls the amount of regularization. The extra term penalizes larger network weights. The procedure is known as weight decay or L2 regularization. Because of the nonlinearity of ANNs, $J(\vec{\theta})$ (or $\tilde{J}(\vec{\theta})$) is a non-convex function. One can still use gradient-based optimizers, but there is no global convergence guarantee. Nevertheless, training algorithms of ANNs are mostly based on using the gradient to descend the cost function to lower values. After initializing all trainable parameters - for example by small random numbers - the gradient of the cost function is used to update the parameters by

\begin{equation}
\vec{\theta}'=\vec{\theta}-\epsilon\vec{\nabla} J,
\end{equation}

where $\epsilon$ is a small parameter called the learning rate that ensures that the change in $\vec{\theta}$ is small. The gradient $\vec{\nabla} J$ of the cost function with respect to $\vec{\theta}$ can effienctly be computed with the backpropagation algorithm. For large training datasets, gradient computation can still be very time consuming. It turns out that the efficiency can be improved by calculating the gradient on small randomized subsets of the training set called minibatches and applying updates to the parameters more often. This procedure is called stochastic gradient descent. Finally, one pass of the full training set is called an epoch.

The use of ANNs for surrogate modelling has advantages and disadvantages. A big advantage is that ANNs can capture the behavior of complicated functions because they can scale to large datasets and also generalize non-locally \cite{Goodfellow2017}. Especially, if a deep network can extract the underlying factors, ANNs are well suited even for high-dimensional problems. The biggest disadvantage is that ANNs mostly act as black boxes. The field of explainable artificial intelligence studies methods to make models like ANNs more explainable and interpretable, but is still in its infancy.

\section{CFD based Data Generation}

\label{sec:CFD based Data Generation}

For the generation of training and test datasets, CFD calculations of supercritical methane flowing inside of straight cooling channel segments are performed. As mentioned in the introduction, many studies have been performed to derive suitable CFD setups, which can reproduce all essential effects influencing the heat transfer. The focus of this paper is to show the feasibility of ANNs to tackle the challenge of numerically efficient heat transfer predictions under the assumption that precise CFD solvers are available for the corresponding problem setting. 

\subsection{CFD Models}

\begin{figure}
	\centering
	\includegraphics[width=421pt]{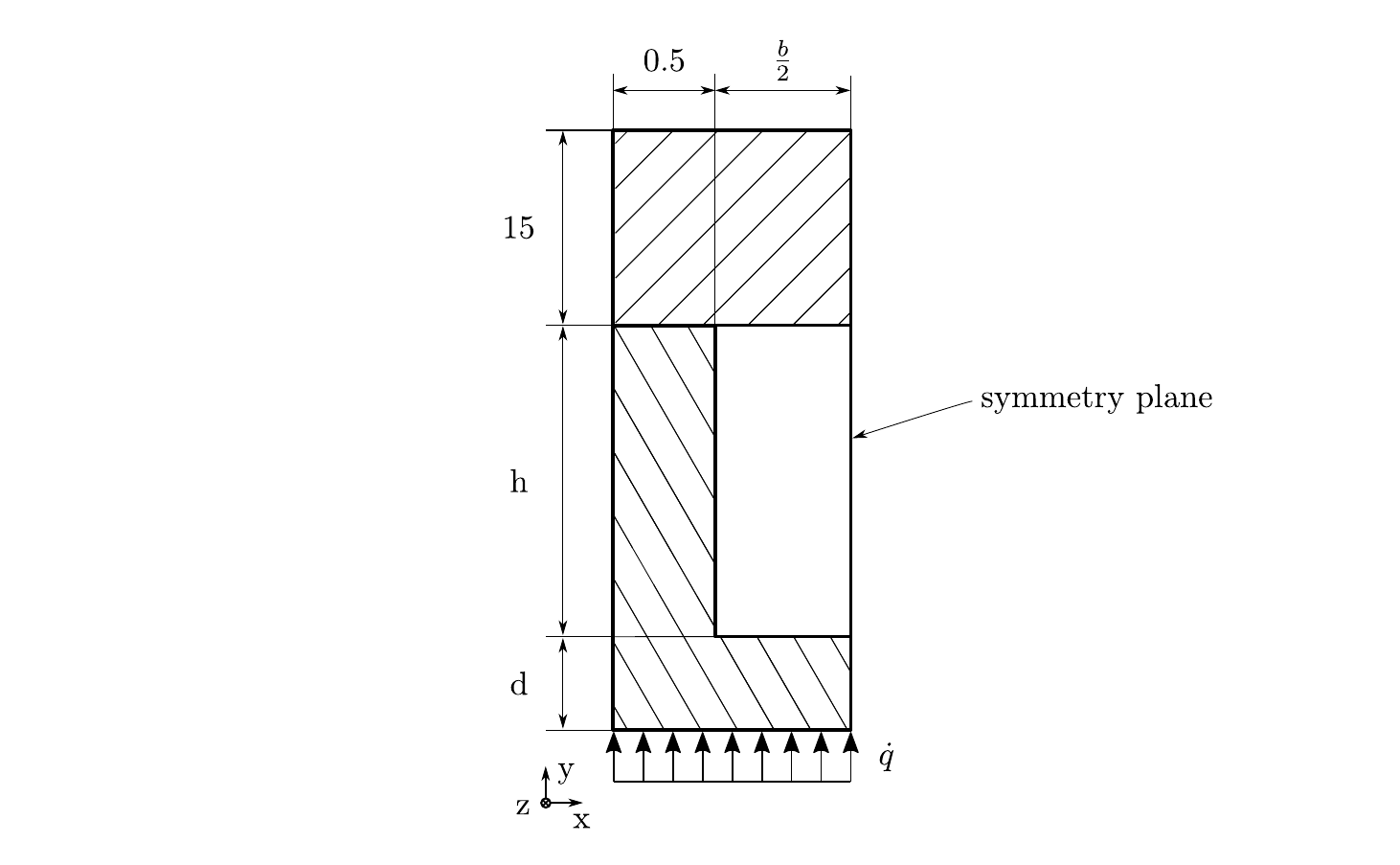}
	\caption[Geometry and boundary conditions of the cooling channels]{Geometry and boundary conditions of the cooling channel (not to scale)}
	\label{fig:geom}
\end{figure}

\begin{figure}
	\centering
	\includegraphics[width=421pt]{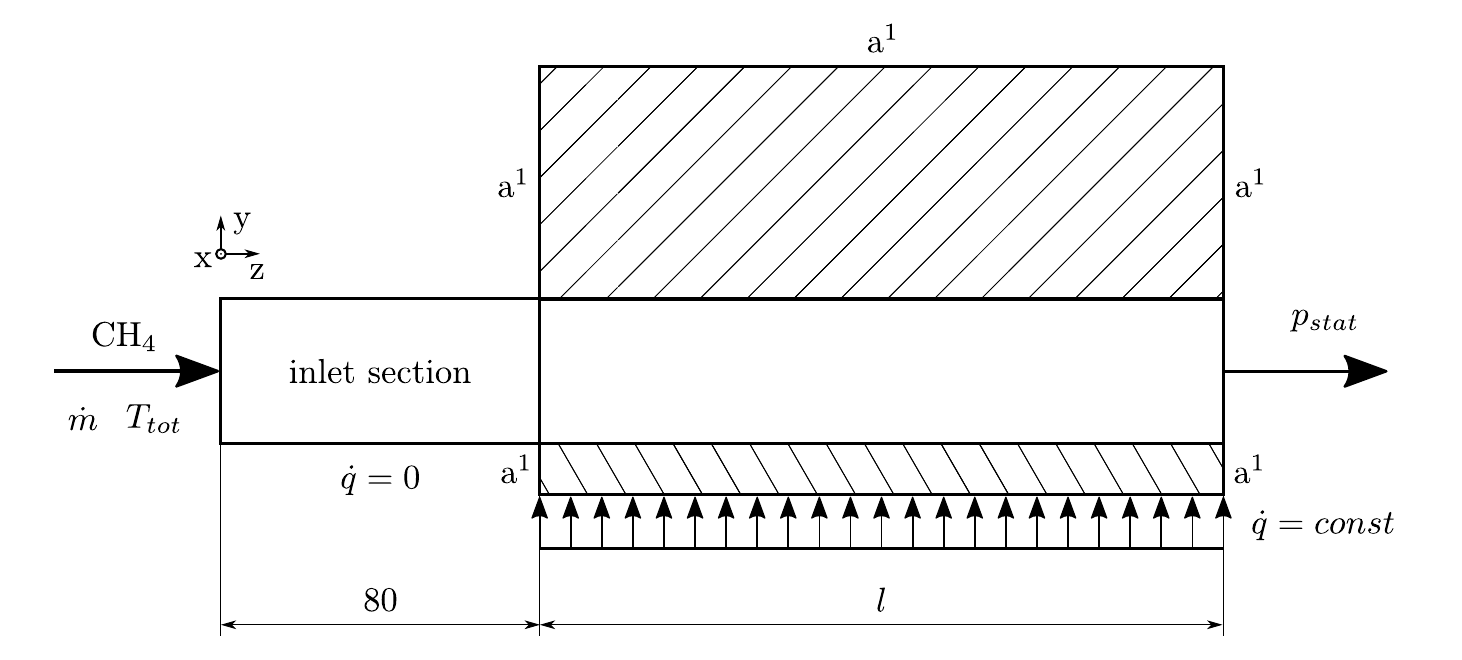}
	\caption[Computational domain with boundary conditions]{Computational domain with boundary conditions (not to scale) -- $^1$walls denoted with "a" are adiabatic}
	\label{fig:boundary}
\end{figure}

The CFD models are generated with standard ANSYS CFX 18.0. The channel flow is modeled as compressible and stationary, while buoyancy and gravitational forces are neglected. As turbulence model the two-equation shear stress transport (SST) model is used which combines the $k-\omega$ turbulence model for the inner region of the boundary layer with the $k-\epsilon$ turbulence model for the free shear flow. The geometry and boundary conditions of the cooling channel model are shown in Fig. \ref{fig:geom} and Fig. \ref{fig:boundary}. Because of symmetry reasons it is sufficient to model one half of the channel. $h$ and $b$ denote channel height and width, while $d$ is used for the chamber wall thickness in the figure. To restrict the independent variables a fin thickness of \SI{1}{\milli \meter} is assumed for all simulations. In stream-wise direction, no heat flux ($\dot{q}=\SI{0}{\watt \per \meter \squared}$) is applied for the first \SI{80}{\milli \meter} of the channel to obtain a fully developed flow and velocity boundary layer. $l$ denotes the channel length and is set to \SI{250}{\milli \meter} for a cross section smaller or equal to \SI{5}{\milli \meter \squared}, while it is increased for channels with a larger cross section to allow the thermal boundary layer to grow further. The channel surface is modeled as a rough wall with different values for the surface roughness and a no-slip condition. A mass flow boundary condition and the coolant total temperature are imposed at the fluid inlet. Furthermore, the static pressure is fixed at the domain outlet and a symmetric flow boundary condition assures no mass or energy fluxes across the symmetry plane. For the solid domain, all faces, except the hot gas wall, are modeled as adiabatic walls. Thermodynamic properties of supercritical methane are evaluated with data from the well known NIST database \cite{Linstrom1997}, which provides data up to \SI{625}{\kelvin}. For higher temperatures, an ideal gas behavior is assumed. The solid domain uses two different material models. Combustion chamber and solid fins consist of a CuCrZr-alloy, which in this case is \SI{99.25}{\percent} copper, \SI{0.62}{\percent} chrome and \SI{0.1}{\percent} zirconium. For the material properties of the alloy the reader is referred to Oschwald et al. \cite{Oschwald2004}. The galvanic layer is assumed to be made of copper. To reduce the influence of axial heat transfer, the thermal conductivity in the stream-wise direction is set to zero for both materials.

The following parameters are varied for data generation: mass flow density $G$, heat flux $\dot{q}$, outlet pressure $p_\text{stat,out}$, inlet temperature $T_\text{stat,in}$, surface roughness $r$, channel area $A$, aspect ratio $\mathit{AR}$ and inner wall thickness $d$. Their upper and lower bounds are chosen so that the data cover the geometrical dimensions and operation conditions of both upper stage and first stage liquid rocket engines with moderate chamber pressure. The outlet pressure ranges between \SI{50}{\bar} and \SI{150}{\bar}, which means that the fluid pressure is always above the critical pressure of methane, and that consequently no boiling or phase change does occur. The fluid inlet temperature varies from \SIrange{120}{400}{\kelvin}. Hence, there are simulations where the coolant temperature crosses the Widom-line and a transition from a liquid-like to a gas-like state takes place. Furthermore, both outlet pressures and inlet temperatures are clustered more narrowly around the critical point to ensure that these critical cases are well represented. To model both smooth and rougher walls, sand-grain roughnesses between \SI{0.2}{\micro \meter} and \SI{15}{\micro \meter} are considered. The channel area varies from \SIrange{1}{10}{\milli \meter \squared} and also different channel aspect ratios (\SIrange{1.0}{9.2}{}) are simulated, because of their impact on heat transfer and maximum wall temperature. For the channel with a cross section of \SI{1}{\milli \meter \squared} only an aspect ratio of \num{1.0} is used to take manufacturing restriction into account. The inner chamber wall thickness varies from \SIrange{0.8}{1.2}{\milli \meter}, which significantly influences the hot gas wall temperature. Generally, higher mass flow densities are considered for high heat fluxes and smoother walls because they result in reasonable wall temperatures and pressure losses. 

For both solid and fluid domains, hexahedral mesh elements are generated with ANSYS ICEM. The first element in the boundary layer has a thickness of \SI{0.1}{\micro \meter} to satisfy a value of y+ < 1. The grid resuolution is \SI{100}{\micro \meter} in stream-wise direction and \SI{35}{\micro \meter} perpendicular to it for the fluid domain. For grid independence, a finer mesh with twice as many elements was analyzed for certain test cases. Because the resulting wall temperatures only change by \SI{2}{\percent} the coarser mesh shows sufficient precision and is therefore used. A converged solution has to fulfill three criteria: All RMS-residuals must be below \num{1e-5}, the conservation equations are well satisfied (solution imbalances below \SI{1}{\percent}) and quantities of interest, such as pressure drop or maximum wall temperature, do not change significantly between two iterations. In total, approximatly \num{20000} CFD simulations of straight cooling channel segments are performed.

\subsection{CFD Results}

\begin{figure}
	\begin{subfigure}[b]{0.5\textwidth}
		\centering
		\includegraphics[width=210pt]{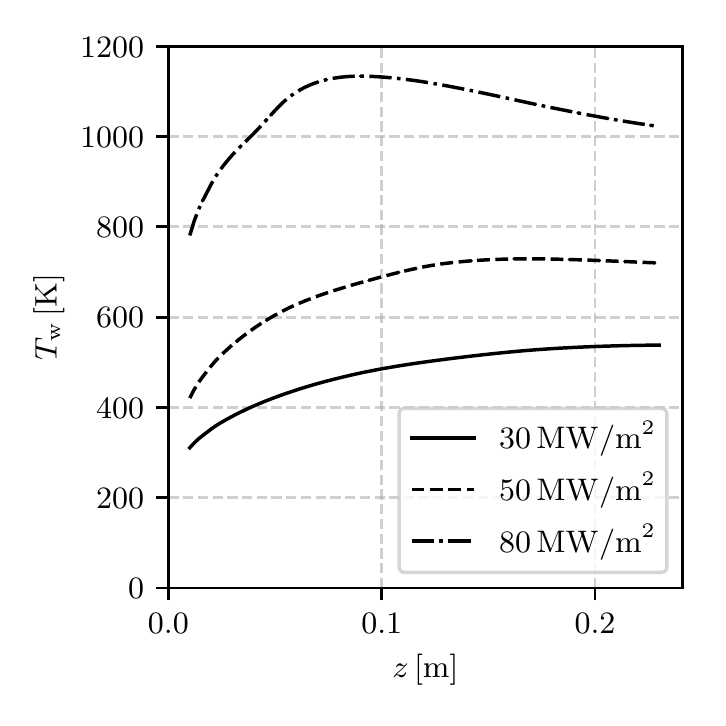}
		\caption{Maximum wall temperature $T_\text{w}$}\label{fig:htd1}
		\label{fig:htd1}
	\end{subfigure}%
	\hfill
	\begin{subfigure}[b]{0.5\textwidth}
		\centering
		\includegraphics[width=210pt]{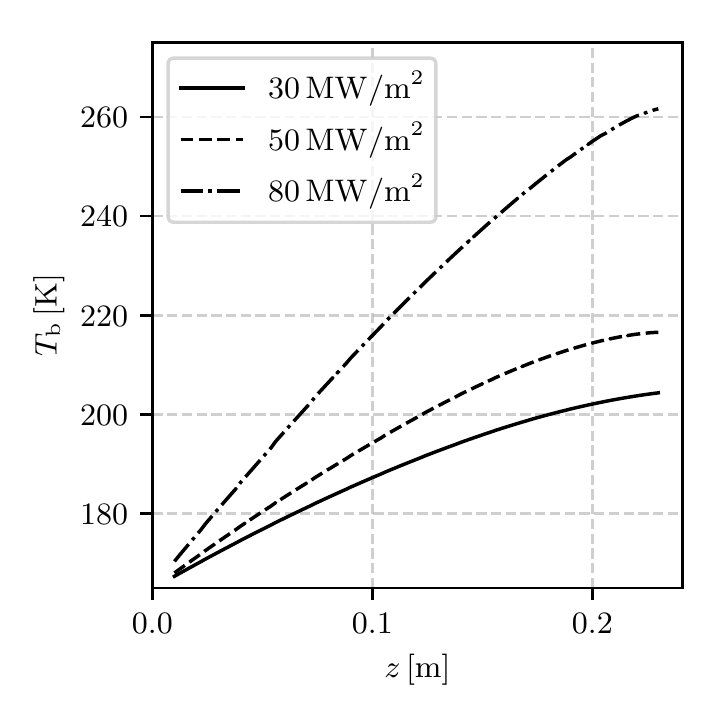}
		\caption{Bulk temperature $T_\text{b}$}\label{fig:htd2}
		\label{fig:htd2}
	\end{subfigure}
	\caption{Wall temperature and bulk temperature for different heat fluxes} \label{fig:htd_cfd}
\end{figure}

ML techniques can only cover effects if they are already present in the training data. Important phenomena, which affect flows in asymmetrically heated cooling channels and the associated heat transfer, are thermal stratification and heat transfer deterioration. Both effects can be observed in the CFD results; e.g. Fig. \ref{fig:htd_cfd} shows the maximum wall temperature and mean bulk temperature along the axial direction of a simulated straight cooling channel for different constant wall heat fluxes. The wall temperature distribution exhibits a peak for higher heat fluxes as a consequence of heat transfer deterioration, while the bulk temperature increases nearly linearly. The influence of the surface roughness is also modelled correctly. Higher roughness levels enlarge the production of turbulence in the boundary layer. Thus, wall temperatures are decreased, but the pressure loss is increased. These implications coincide with the CFD results. Overall, it can be concluded that the relevant consequences of thermal stratification, heat transfer deterioration, and surface roughness are represented in the generated data. 

\subsection{Data Reduction}

\begin{figure}
	\centering
	\includegraphics[width=420pt]{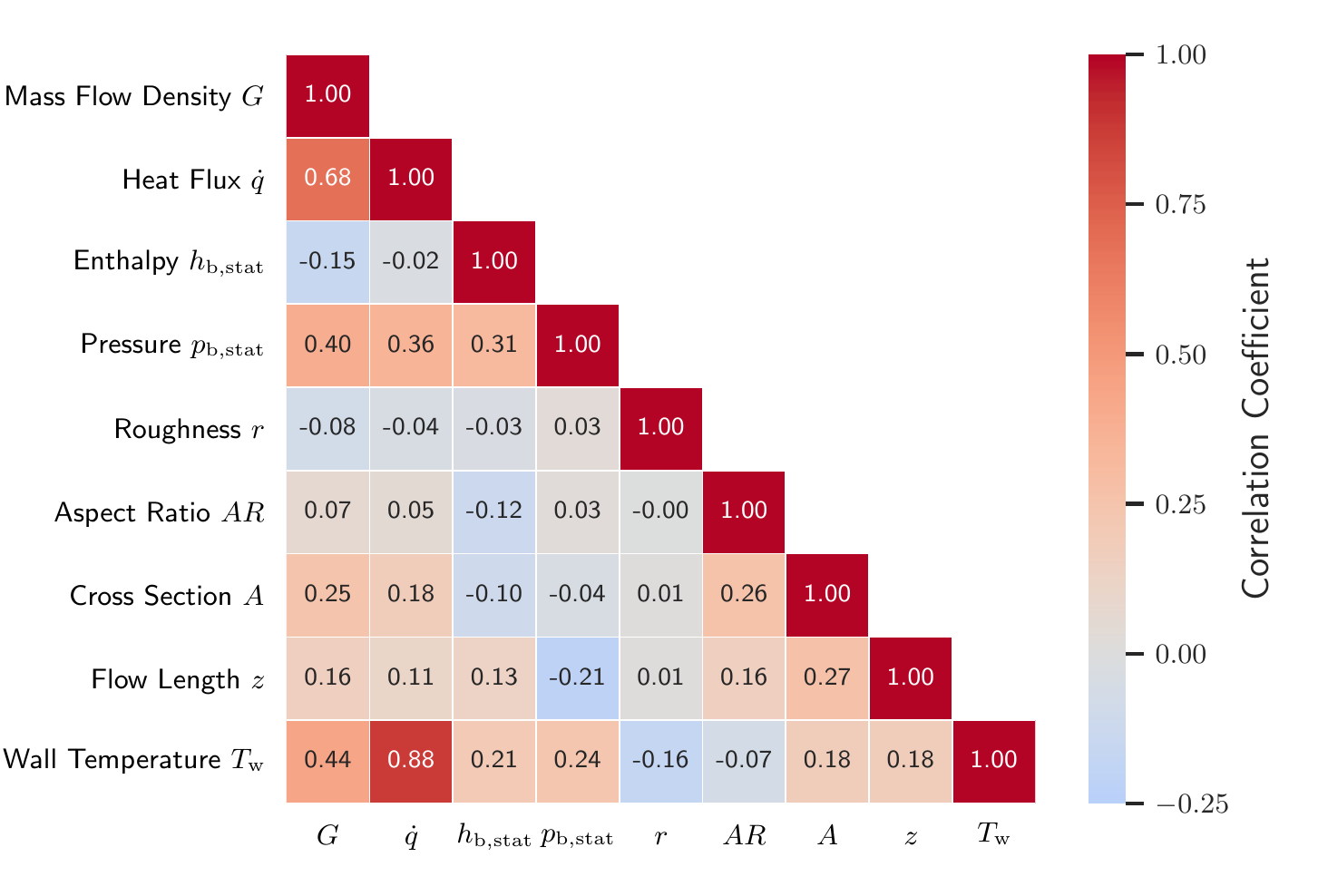}
	\caption{Correlation matrix}
	\label{fig:corr}
\end{figure}

Only a reduced amount of the CFD results is used for training the ANN. First, only the values of bulk properties are utilized for the fluid description. Bulk properties are calculated as mass-flow averaged quantities across the channel cross section. Although most information contained in the two-dimensional distribution of fluid quantities is lost, it is hoped that the impact will be reflected in the correlations of the bulk variables. Second, at each cross section the temperature distribution of the solid part is reduced to the values of the mean wall temperatures and the maximum temperature wall temperature at the hot gas side. Nevertheless, it would be interesting to check how far the accuracy of a data-driven model can be increased by using a more complete description of the fluid and solid states. Third, these variables are only evaluated every \SI{2}{\milli \meter} in stream-wise direction and saved in a table-like file structure. Each data point is extended by the associated geometric information, such as cross section area, aspect ratio, flow length as well as boundary conditions like heat flux and surface roughness. The flow length is used to include boundary layer effects on the heat transfer.

\begin{figure}
	\centering
	\includegraphics[width=420pt]{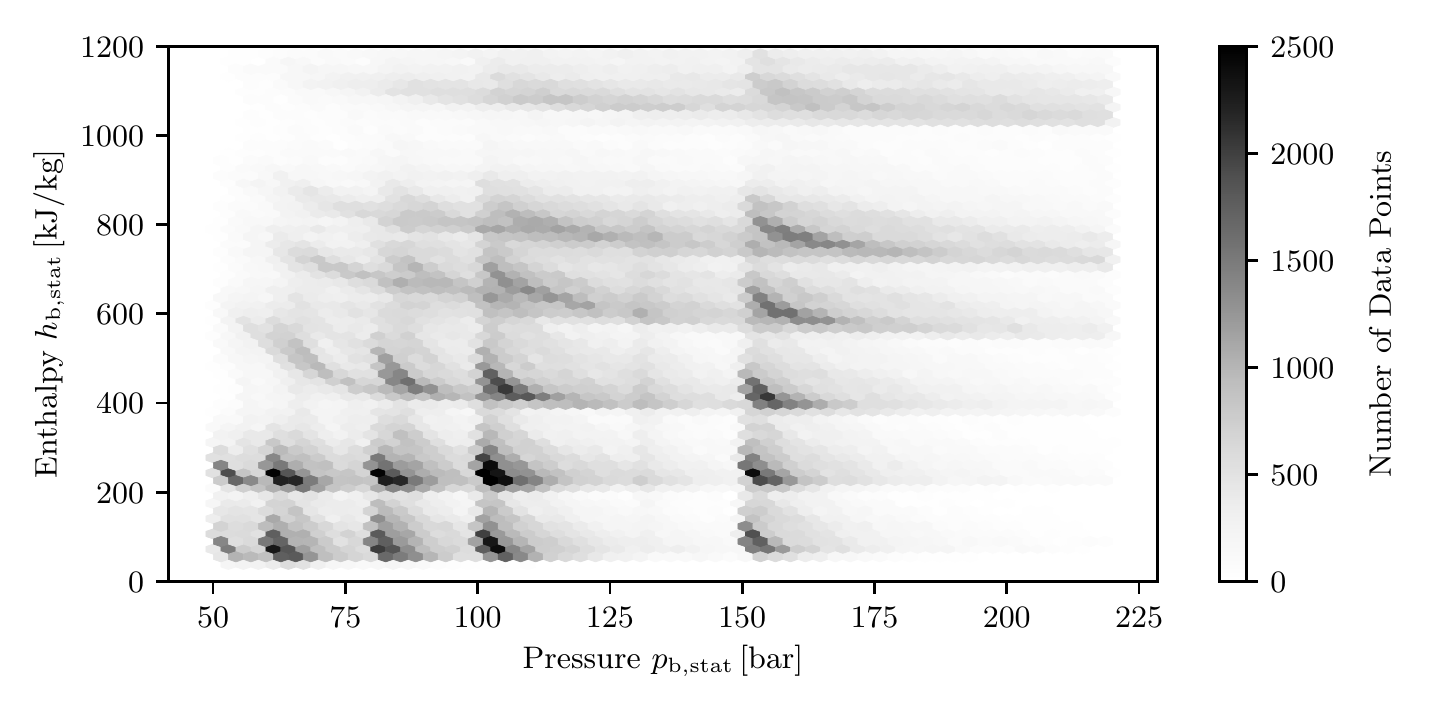}
	\caption{Data distribution}
	\label{fig:raw_distribution}
\end{figure}

After data generation, it is always recommended to study the content and distribution of the data. First, a correlation matrix can be used to visualize the correlations between multiple variables. Figure \ref{fig:corr} shows the correlation matrix for certain variables, where each entry visualizes the value of the corresponding Pearson correlation coefficient. It is important to note that Pearson correlation only describes the strength of linear relationships and does not imply causation. For example, the correlation coefficient between wall temperature and heat flux is \num{0.88} and indicates a strong positive relationship. Whereas the correlation between wall temperature and surface roughness is negative as expected. One can see that physically reasonable relations are still represented in the reduced data. Second, the data is not uniformly distributed and there are regions where the data is very sparse or where no data points are present at all. Figure \ref{fig:raw_distribution} exemplary shows the distribution with respect to enthalpy and pressure. As a result of the data generation process with its manually chosen boundary conditions, there are regions with higher and lower data density. A so-called covariate shift refers to a situation where the distribution of input variables is different in the data available for training and the data one expects to use as input in the future \cite{Sugiyama2007}. This needs to be taken into account because the ANN should also produce good predictions there.

\section{Artificial Neural Network for Wall Temperature Prediction}
\label{sec:Artificial Neural Network for Wall Temperature Prediction}

An important problem is the prediction of the maximum temperature for each section of the combustion chamber wall given a certain cooling channel design and suitable boundary conditions. The maximum temperature is a critical parameter, because it directly determines the fatigue life of the chamber and is therefore a crucial constraint for design considerations \cite{Waxenegger2017}. The main driver for the temperature is the heat transfer from the cooling channel to the coolant. Hence, the prediction can only be successful if the implicit heat transfer modelling takes the underlying mechanisms correctly into account. Put differently, this means that an accurate wall temperature prediction implies a proper reduced order modelling of the relevant heat transfer. 

\begin{figure}
	\centering
	\includegraphics[width=420pt]{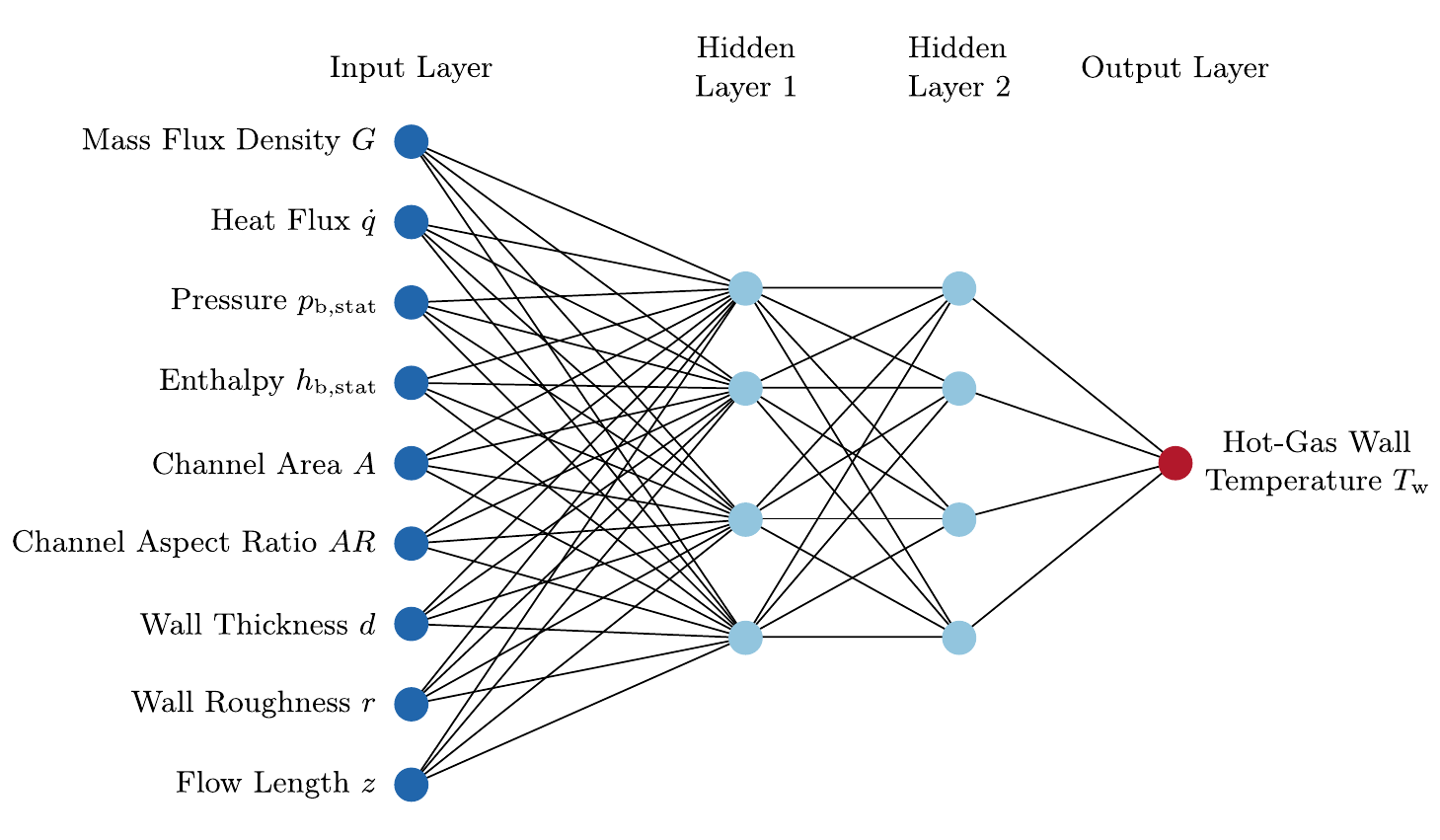}
	\caption{Exemplary network architecture with 2 fully connected hidden layers}
	\label{fig:nn_without_flowlength}
\end{figure}

\subsection{Network Architecture and Hyperparameter Optimization}

A fully connected, feedforward network is proposed for the wall temperature prediction. The term fully-connected means that every neuron of one layer is linked with all neurons of the next layer. Figure \ref{fig:nn_without_flowlength} shows an exemplary model with two hidden layers, four neurons per hidden layer and all input parameters. The optimal number of hidden layers and neurons depends on the specific problem and data respectively. To find the best network architecture and training parameters, one needs to split the available data into suitable training and validation sets. Therefore, \SI{90}{\percent} of the data points is randomly selected for training and the rest is held back for validation. However, under a covariate shift, data points should be weighted according to their so-called importance, which can be calculated by kernel density estimations, when calculating error measures for training and validation. For further details the reader is referred to Sugiyma et al. \cite{Sugiyama2007}. 

Given training and validation data, classical grid search and random search can be used to determine the optimal parameters. Bergstra and Bengio \cite{Bergstra2012} showed that a random search algorithm performs as well as grid search but with less computational cost. The proposed ANN uses ReLUs for the activation functions of the hidden neurons and a linear unit is employed for the continuous output. During training, the weight and bias update is calculated with the ADAM optimizer, which is an extension of the classic stochastic gradient descent algorithm \cite{Kingma2014}. For faster and more robust learning, all inputs are automatically scaled and standardized with the StandardScaler from Scikit-learn \cite{Pedregosa2011}. The cost function is given by a mean squared error term plus an extra term for L2 regularization, as in Eq. \ref{eq:cost_regularized}. The model is generated and trained with KERAS, which is an open-source ANN library written in Python \cite{Keras}. Random search of 500 different hyperparameter combinations and network architectures leads to the following optimal model:

\begin{itemize}
\item 4 hidden layers
\item 408 neurons per hidden layer
\item L2 regularization with $\alpha=0.1$
\item minibatch size of 4096
\item 150 epochs
\end{itemize}

\noindent The training takes about 15 minutes on a Nvidia Quadro P4000 GPU.

\subsection{Results and Visualization}

\begin{figure}[h]
	\centering
	\includegraphics[width=420pt]{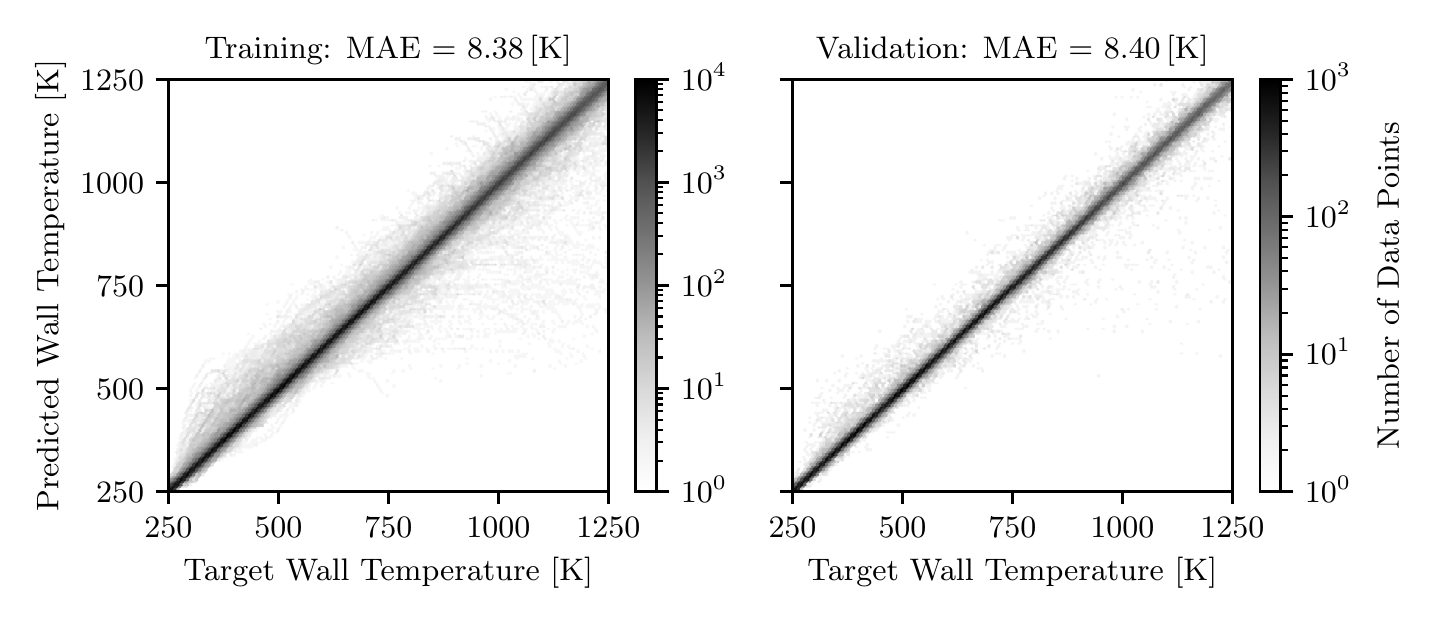}
	\caption{Training and validation results for the proposed model}
	\label{fig:target_vs_expected}
\end{figure}

Figure \ref{fig:target_vs_expected} compares predicted and target values for the wall temperature. One can see that the proposed network achieves a convincing precision. The mean absolute error (MAE) of the wall temperature prediction is \SI{8.38}{\kelvin} on the training set and \SI{8.40}{\kelvin} on the validation set with a standard deviation of \SI{17.7}{\kelvin} and \SI{18.5}{\kelvin}, respectively. The reason for the smaller error on the training data is the fact that the training data was directly used to optimize the model's weights, but the performance on the validation data is still impressive. One can conclude that a suitable selection of input variables are chosen to predict the maximum wall temperature with high precision. Furthermore, the amount of data samples is sufficient to train the network. Hence, one would conclude that the network has generalized well and does not overfit. 

Nevertheless, there is still the risk of overfitting, especially because of the empty regions in the input space, which are also present in the validation set. To evaluate the quality of the ANN model, it is necessary to study the performance on an independent test set with yet unseen data. For this purpose, 25 further CFD calculations for 5 different channel geometries are made and the resulting maximum wall temperatures are compared with the predictions of the ANN. The input parameters of 6 exemplary simulations are presented in Table \ref{fig:test_cases}. To include various engine sizes and operation conditions, the boundary conditions are varied in a wide range leading to lower but also higher wall temperatures. As both channel geometries and operational conditions differ from those of the training and validation data, the test set is an unbiased and independent performance measure for the ANN. The reader is referred to the appendix for a detailed overview of the train and test data distributions.

\begin{figure}[p]
	\centering
	\includegraphics[width=420pt]{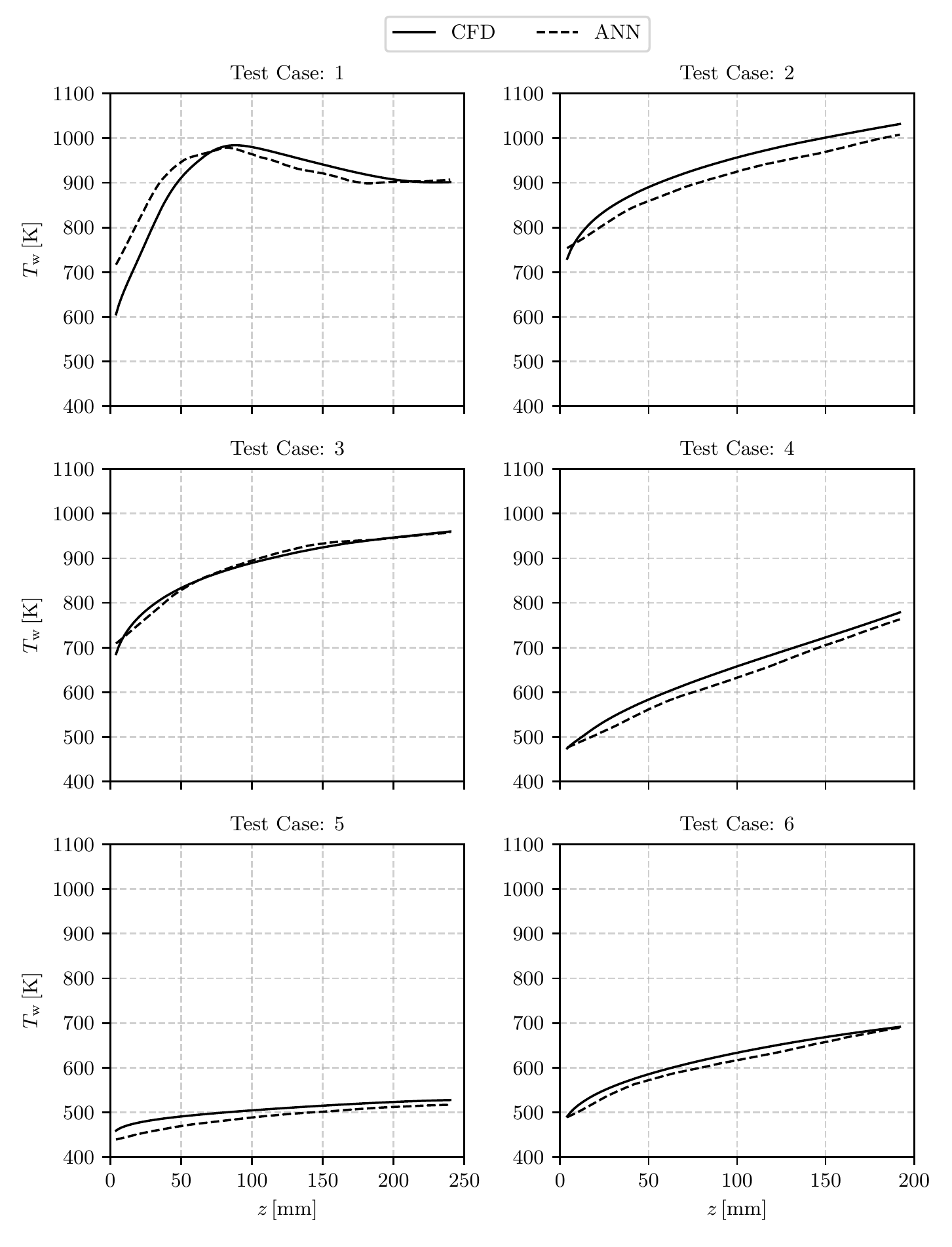}
	\caption{Wall temperature prediction for various different design points, which are representative for different operation conditions and cooling channel geometrie (see table \ref{fig:test_cases})}
	\label{fig:test_set_cfd_for_h_and_p}
\end{figure}

\begin{table}[]
	\centering
	\sisetup{per-mode=fraction}
	\begin{tabular}{@{}crrrrrrrr@{}}
		\toprule
		\multicolumn{1}{c}{{Test Case}}  & \multicolumn{1}{c}{{$T_\text{in}$}} & \multicolumn{1}{c}{$p_\text{out}$}  & \multicolumn{1}{c}{$\dot{q}$} & \multicolumn{1}{c}{$G$} & \multicolumn{1}{c}{$A$} & \multicolumn{1}{c}{$\mathit{AR}$} & \multicolumn{1}{c}{${d}$} & \multicolumn{1}{c}{$r$} \vspace{0.05cm}\\
		
		\multicolumn{1}{c}{[--]} & \multicolumn{1}{c}{[\si{\kelvin}]} & \multicolumn{1}{c}{[\si{bar}]} & \multicolumn{1}{c}{[\si{\mega \watt \per \meter \squared}]} & \multicolumn{1}{c}{[\si{\kilogram \per \meter \squared \per \second}]} & \multicolumn{1}{c}{[\si{\milli \meter \squared}]}  & \multicolumn{1}{c}{[--]} & \multicolumn{1}{c}{[\si{\milli \meter}]} & \multicolumn{1}{c}{[\si{\micro \meter}]}  \\
		\midrule
		1 &	140	&	80	&	49	&	\num{11700}	 &	1.9	&	2.0	&	0.83	&	2.1	\\
		2 &	131	&	217	&	81	&	\num{11700}	 &	4.1	&	4.1	&	0.90	&	3.0	\\
		3 &	173	&	129	&	57	&	\num{23900}	 &	7.4	&	3.7	&	1.14	&	3.0  \\
		4 &	127	&	57	&	55	&	\num{26000}	 &	6.0	&	7.5	&	0.96	&	14.2   \\
		5 &	290	&	51	&	14	&	\num{10100}	 &	7.4	&	3.7	&	1.14	&	1.7		\\
		6 &	148	&	174	&	37	&	\num{13200}	 &	3.2	&	2.3	&	1.07	&	6.4 	\\
		\bottomrule
	\end{tabular}
	\caption{Exemplary boundary conditions for the test dataset}
	\label{fig:test_cases}
\end{table}

Figure \ref{fig:test_set_cfd_for_h_and_p} shows the maximum wall temperature as a function of the axial length for both the CFD simulation and the ANN. The MAE is \SI{16.0}{\kelvin} with a standard deviation of \SI{12.0}{\kelvin}. Overall, the ANN shows convincing performance for all wall temperature regimes. Wall temperatures up to \SI{1000}{\kelvin} and as low as \SI{500}{\kelvin} are predicted with minimal error. By using the flow length, the trained network is also able to reproduce the non-linear evolution in stream-wise direction. Finally, the effect of heat transfer deterioration is learned as test case \num{1} shows. It can be concluded that the ANN has captured the essential underlying factors. It successfully predicts the maximum wall temperatures, even for regions in the input space where no training or validation data points are present.

\begin{figure}[h]
	\centering
	\includegraphics[width=466pt]{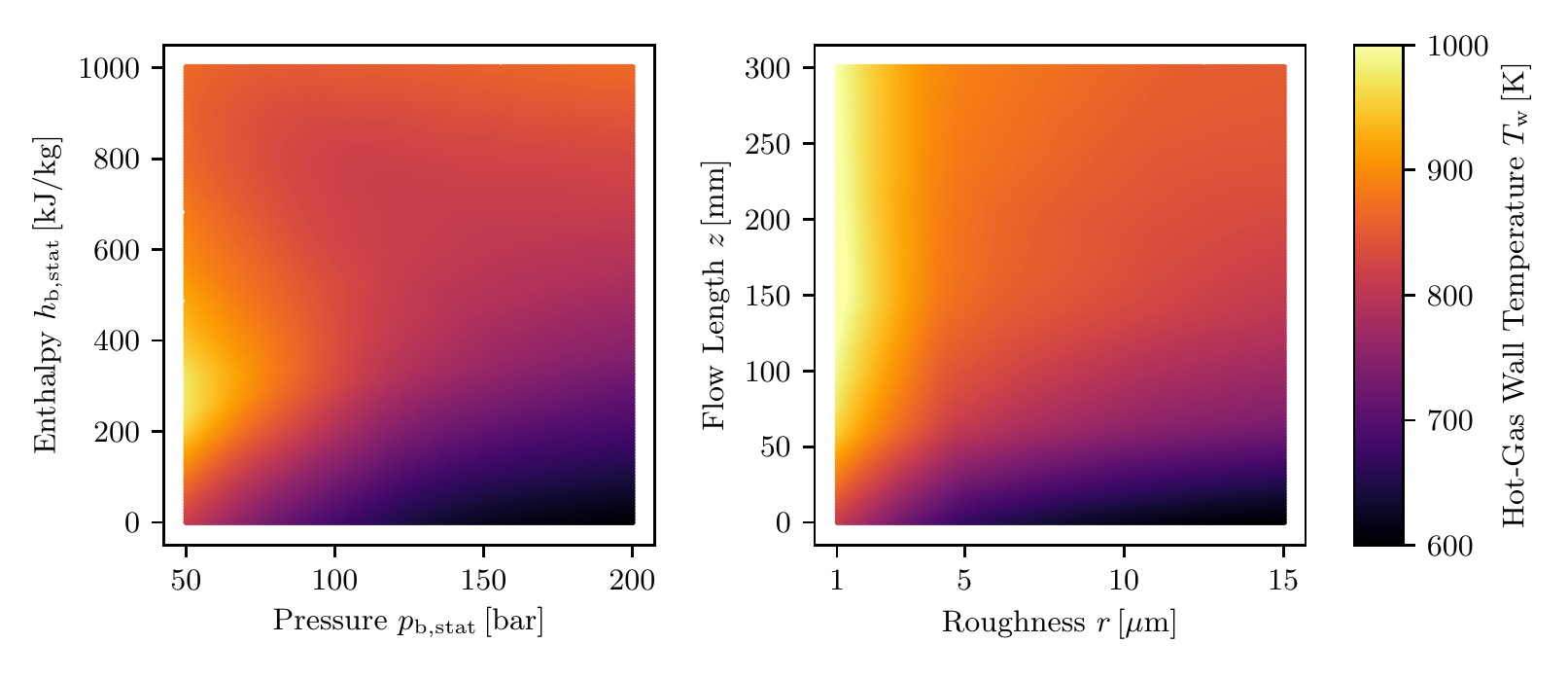}
	\caption{Exemplary heat maps for the trained network}
	\label{fig:response_surface}
\end{figure}

Finally, it is often useful to visualize the prediction of the model for a certain range of input values. Directly observing the output helps to decide whether a model has learned the fundamental  underlying factors of the given task or if it merely memorizes the training data. Additionally, it is important to visualize how the model performs in between of the given input data. One can employ so-called heat maps, where two inputs are parametrically changed while all other parameters are kept the same. The output is then plotted in a two-dimensional scatter plot. Heat maps can be used to identify possible problems in terms of overfitting. For example, further investigations would be necessary if there are regions with strong unexpected discontinuities.
Figure \ref{fig:response_surface} illustrates the effect of varying coolant temperature and pressure as well as the influence of channel rougness and flow length. In terms of physical interpretation, the response seems reasonable. In general, a lower coolant bulk enthalpy or a higher bulk pressure lead to lower wall temperatures for a given heat flux because of changes in the transport properties of the coolant. Furthermore, the wall temperature builds up excessively close to the critical point. Higher roughness levels enlarge the production of turbulence in the boundary layer, thus decrease the wall temperature. The flow lenghts reflects the influence of boundary layer growth. Finally, the wall temperature prediction changes smoothly without unphysical discontinuities.

\section{Reduced Order Model for Cooling Channel Flow}
\label{sec:Reduced Order Model for Cooling Channel Flow}

In addition to forecasting maximum wall temperatures, the prediction of critical variables such as pressure loss and heating of the coolant is essential for regenerative cooling design. If CFD calculations are not suitable due to their high computational cost, further reduced order models are required to calculate the stream-wise development of thermodynamic properties like pressure and enthalpy, while the ANN is used to predict the wall temperature.

\subsection{Pressure Drop Model}

The Darcy-Weisbach equation can be used to to estimate the pressure loss along the cooling channels. The pressure loss in a channel segment of length $\Delta z$ is given by

\begin{equation}
\Delta p = \frac{1}{2}f\rho_\text{b} v_\text{b}^2\frac{\Delta z}{D_{h}},
\end{equation}
\label{eq:pressure_drop}

\noindent where $f$ is the so called friction factor, $\rho_\text{b}$ the bulk density of the coolant, $v_\text{b}$ the bulk flow velocity and $D_{h}$ the hydraulic diameter of the channel. The friction factor $f$ can be calculated by means of a simple empirical correlation valid for laminar, turbulent and transient flow \cite{Churchill1977}:

\begin{equation}
f = 8\left[\left(\frac{8}{Re}\right)^{12}+\frac{1}{(A+B)^{1.5}} \right]^{1/12}
\end{equation}
\label{eq:friction_factor}

\noindent with

\begin{equation}
A =\left[2.457 \ln\left(\frac{1}{\left(\frac{7}{Re}\right)^{0.9}+0.27\left(\frac{r}{D_{h}}\right)}\right)\right]^{16}\quad\text{and}\quad B = \left(\frac{37530}{Re}\right)^{16},
\end{equation}
\label{eq:const_A_B}

\noindent where $Re$ denotes the local Reynolds number and $r$ the surface roughness.

\subsection{Enthalpy Increase Model}

Conservation of energy calculates the change of the specific total enthalpy of the fluid over a channel section of length $\Delta z$: 

\begin{equation}
h_{b,tot}(z+\Delta z) = h_{b,tot}(z)+\frac{\dot{Q}(z,\Delta z)}{\dot{m}}\quad\text{with}\quad h_{b,tot}(z)  = h_\text{b,stat}(z)+\frac{1}{2}v_\text{b}(z)^2,
\end{equation}
\label{eq:enthalpy_increase}

where $z$  is the stream-wise coordinate, $h_\text{b,stat}$ the specific bulk enthalpy of the fluid, $v_\text{b}$ the bulk flow velocity, $\dot{m}$ the mass flow rate and $\dot{Q}$ the overall heat flow rate in the channel segment.

\subsection{Comparison with CFD}

\begin{figure}
	\centering
	\includegraphics[width=420pt]{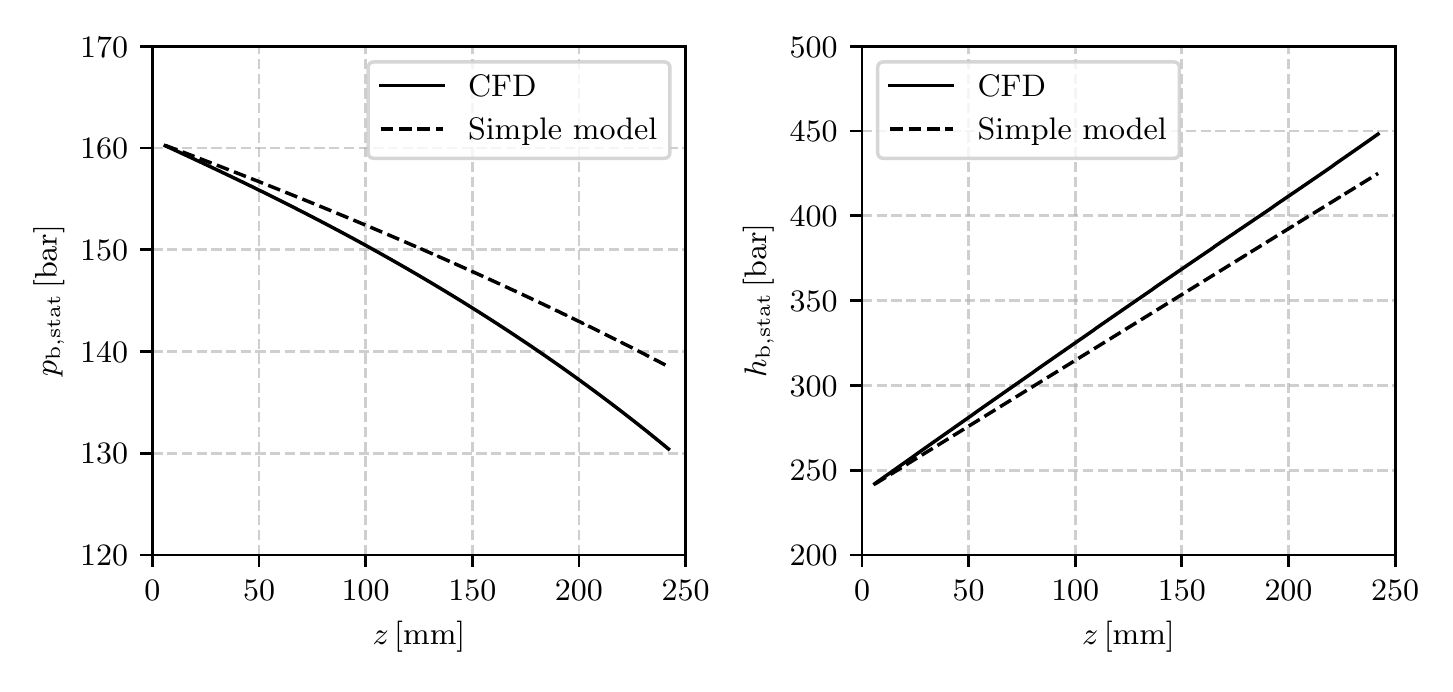}
	\caption{Comparison between simple one-dimensional models and CFD data for bulk enthalpy and bulk pressure for an exemplary test case}
	\label{fig:test_case_ab_1_cfd_for_h_and_p}
\end{figure}

\begin{figure}[p]
	\centering
	\includegraphics[width=420pt]{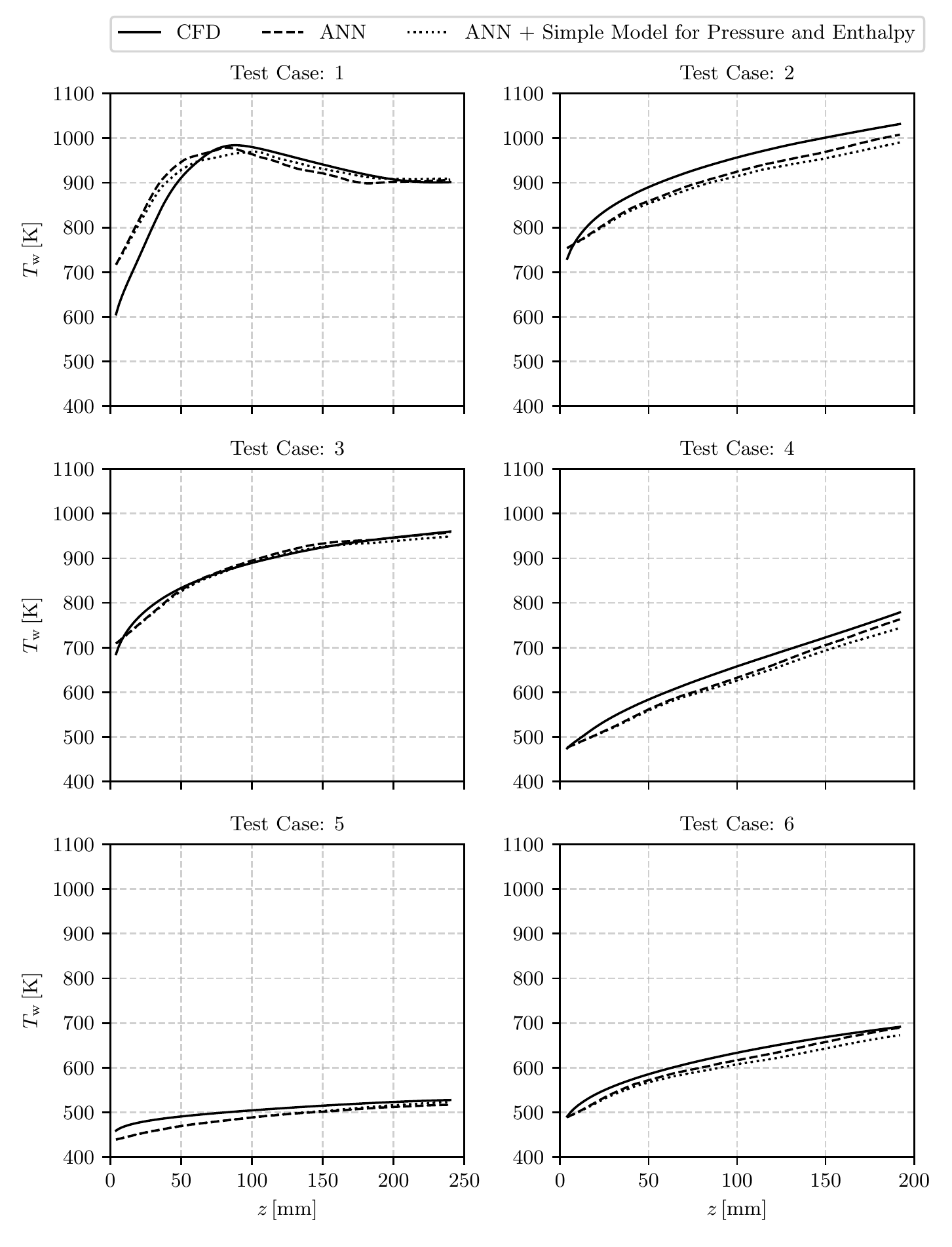}
	\caption{Comparison of wall temperatures for CFD, the reduced order model and a hybrid model that uses pressure and enthalpy from CFD and the ANN for wall temperature prediction}
	\label{fig:test_set_surrogate_model_for_h_and_p}
\end{figure}

If one adds a mass continuity equation and a suitable equation of state (or uses the NIST database), one obtains a complete reduced order model for supercritical methane flowing in a rocket engine cooling channel. The predictions of the reduced order model can be compared with the results of a full CFD calculation. First, Fig. \ref{fig:test_case_ab_1_cfd_for_h_and_p} shows the evolution of the bulk pressure and bulk enthalpy for an exemplary test case. Although error propagation increases the error in stream-wise direction, the simple models produce results with sufficient accuracy. The mean absolute percentage error for enthalpy and pressure on the entire test data set is \SI{4.3}{\percent} and \SI{4.2}{\percent}, respectively. Thus, these models can be used to calculate pressure and enthalpy along a channel, which, in turn, the ANN can use as input for the wall temperature prediction. Figure \ref{fig:test_set_surrogate_model_for_h_and_p} shows the wall temperature prediction for the proposed network using the reduced order models for input generation. The error only marginally increases from \SI{16.0}{\kelvin} to \SI{19.6}{\kelvin} when using the reduced order model for pressure and enthalpy calculation. In summary, the proposed reduced order model is able to predict the evolution of the bulk pressure, the bulk enthalpy, and the resulting maximum wall temperature for low heat fluxes in the range of \SI{10}{\mega \watt \per \meter \squared} (test case 5), medium heat fluxes (test case 1 and 6), as well as very high heat fluxes up to \SI{80}{\mega \watt \per \meter \squared} (test case 2, 3 and 4), which can occur in the nozzle throat of a liquid rocket engine.

\subsection{Performance Assessment}

Although ANNs require a time-intensive training phase, the predictive speed is very high, because the network is just a composite function that multiplies matrices and vectors together. Additionally, the numerical effort does not depended on the actual value of the inputs (e.g. channel area), whereas CFD simulations need increasingly more time with larger model sizes and thus higher number of mesh elements. On a computer with an Intel Xeon Gold 6140 CPU the CFD calculation of one straight channel segment takes up to 1 hour depending on the channel cross section, while the reduced order model delivers the result after \SI{0.6}{\second}. This comparison shows the great potential of data-driven surrogate models for design space explorations and optimization loops. 

\section{Conclusion and Outlook}

\label{sec:Conclusion and Outlook}

In this paper, an ANN was successfully trained to predict the maximum wall temperature for each cylindrical section of a rocket combustion chamber wall given a regenerative cooling design using supercritical methane and suitable boundary conditions. The network was trained on data generated by CFD simulations of straight cooling channel segments. The ANN predicts the wall temperature for previously unseen test cases, including different channel geometries and operation conditions, with an MAE of \SI{16.0}{\kelvin}. Furthermore, the prediction of an entire channel segment takes only \SI{0.6}{\second}, which is at least $10^3$ times faster than comparable three-dimensional CFD simulations. Thus, this numerically efficient method constitutes a convincing building block of a reduced order model for supercritical methane flowing in rocket engine cooling channels. It is also shown which further reduced order models can be added to obtain a suitable description for cooling channel design considerations. The presented methodology can be used to generate predictions with a precision similar to full CFD calculations and after training the answer only takes a fraction of the computation time of a comparable CFD simulation. Therefore, it is well suited for optimization loops and as a component of system analysis tools.

However, ANNs have disadvantages too. On the one hand, there are disadvantages that all data-driven surrogate models share. The data sample selection determines the reachable accuracy. First, if the underlying data is wrong, the resulting model will be wrong as well. For the described methodology this means that it only works if there is a CFD code available which can model all relevant effects, e.g. heat transfer deterioration or the correct influence of different surface roughness levels. Second, depending on the complexity of the problem, the construction of a precise approximation model can require a huge number of data samples. This data generation can get computationally very expensive. One challenge of surrogate modeling is the generation of a model that is as accurate as needed, using as few simulation evaluations as possible. It would be interesting to study how the additional use of experimental data could improve the situation. On the other hand, there are disadvantages that are typical for ANNs. ANNs are not able to extrapolate, but only provide reliable predictions within the region of the input space that is populated with training points. It is important to take this into account when using ANN based models for design space exploration or optimization. Furthermore, due to the high number of parameters, these algorithms often lack a deeper understanding of the fundamental physics. Thus, domain knowledge and the understanding of physical processes is still crucially important to evaluate and justify the prediction of data-driven algorithms.

The present work can be improved in many directions. Clearly, the data generation process is not optimal. The density of the data points is too far from being uniform in the input space of interest. In future research, an optimization of the data generation should be studied. Building on this, the question should be examined how much data is needed to reach a certain accuracy. A different choice of input parameters may increase the precision. Parameters like the boundary layer thickness were not explicitly used in the current model.  A further extension should study the consideration of curvature effects. It is well known that centrifugal forces induce recirculation phenomena in the flow which influence the heat transfer and should not be neglected. Eventually, the performance of ANNs should be compared with other types of surrogate models for the task of wall temperature prediction and heat transfer modelling respectively. Overall, it is hoped that the current work will serve as a basis for future studies regarding the application of ANNs in the field of rocket engine design.

\section*{Appendix}
\label{sec:appendix}

Table \ref{tab:properties_train} and Table \ref{tab:properties_test} give an overview of mean value, standard deviation and different percentiles of most relevant thermodynamic properties of the coolant, channel geometries and the resulting wall temperature at the hot-gas wall for the training and test datasets.

\begin{table}[h]
	\centering
	\sisetup{per-mode=fraction}
	\begin{tabular}{@{}rrrrrrrrrrrr@{}}
		\toprule
		& \multicolumn{1}{c}{{$T_\text{b}$}}   & \multicolumn{1}{c}{{$h_\text{b}$}} & \multicolumn{1}{c}{{$p_\text{b}$}} & \multicolumn{1}{c}{{$v_\text{b}$}} & \multicolumn{1}{c}{{$G$}} & \multicolumn{1}{c}{{$\dot{q}$}} & \multicolumn{1}{c}{{$r$}} & \multicolumn{1}{c}{{$A$}} & \multicolumn{1}{c}{{$\mathit{AR}$}} & \multicolumn{1}{c}{$d$} & \multicolumn{1}{c}{{$T_\text{w}$}} \vspace{0.05cm} \\
		& \multicolumn{1}{c}{{[\si{\kelvin}]}} & \multicolumn{1}{c}{{[\si{\kilo \joule \per \kilogram}]}} & \multicolumn{1}{c}{{[\si{bar}]}} & \multicolumn{1}{c}{{[\si{\meter \per \second}]}} & \multicolumn{1}{c}{{[\si{\kilogram \per \second \per \meter \squared}]}} & \multicolumn{1}{c}{{[\si{\mega \watt \per \meter \squared}]}}  & \multicolumn{1}{c}{{[\si{\micro \meter}]}} & \multicolumn{1}{c}{{[\si{\milli \meter \squared}]}} & \multicolumn{1}{c}{{[--]}} & \multicolumn{1}{c}{{[\si{\milli \meter}]}} & \multicolumn{1}{c}{{[\si{\kelvin}]}} \\
		\midrule
		{Mean} 				& 251  & 566	& 125 	& 126	& \num{18483} 	& 36  & 6.9 & 6.7 & 4.4 & 1.0 & 669 \\
		{Std}  				& 84  & 317		& 42 	& 78	& \num{8078} 	& 24  & 6.1 & 3.2 & 3.1  & 0.1 & 302 \\
		\midrule
		\SI{1}{\percent}  	& 123  & 56 	& 53 	& 18	& \num{3027} 	& 9  & 0.2 & 1.0 & 1.0 & 0.8 &	230 \\
		\SI{25}{\percent} 	& 183  & 279 	& 90 	& 64	& \num{12500} 	& 10  & 1.0 & 5.0 & 1.7  & 1.0 & 426 \\
		\SI{50}{\percent} 	& 240  & 572 	& 119 	& 109	& \num{17500} 	& 30  & 5.0 & 5.0 & 3.5 & 1.0 & 620 \\
		\SI{75}{\percent}  	& 302  & 790 	& 158 	& 174	& \num{25000} 	& 50  & 15.0 & 10.0 & 9.2 & 1.0 & 854 \\
		\SI{99}{\percent}  	& 433  & 1175 	& 215 	& 357	& \num{35000} 	& 80 & 20.0 & 10.0 & 9.2 & 1.2 & 1482 \\
		\bottomrule
	\end{tabular}
	\caption{Mean value, standard deviation and percentiles of the training data}
	\label{tab:properties_train}
\end{table}

\begin{table}[h]
	\centering
	\sisetup{per-mode=fraction}
	\begin{tabular}{@{}rrrrrrrrrrrr@{}}
		\toprule
		& \multicolumn{1}{c}{{$T_\text{b}$}}   & \multicolumn{1}{c}{{$h_\text{b}$}} & \multicolumn{1}{c}{{$p_\text{b}$}} & \multicolumn{1}{c}{{$v_\text{b}$}} & \multicolumn{1}{c}{{$G$}} & \multicolumn{1}{c}{{$\dot{q}$}} & \multicolumn{1}{c}{{$r$}} & \multicolumn{1}{c}{{$A$}} & \multicolumn{1}{c}{{$\mathit{AR}$}} & \multicolumn{1}{c}{${d}$} & \multicolumn{1}{c}{{$T_\text{w}$}} \vspace{0.05cm} \\
		& \multicolumn{1}{c}{{[\si{\kelvin}]}} & \multicolumn{1}{c}{{[\si{\kilo \joule \per \kilogram}]}} & \multicolumn{1}{c}{{[\si{bar}]}} & \multicolumn{1}{c}{{[\si{\meter \per \second}]}} & \multicolumn{1}{c}{{[\si{\kilogram \per \second \per \meter \squared}]}} & \multicolumn{1}{c}{{[\si{\mega \watt \per \meter \squared}]}}  & \multicolumn{1}{c}{{[\si{\micro \meter}]}} & \multicolumn{1}{c}{{[\si{\milli \meter \squared}]}} & \multicolumn{1}{c}{{[--]}} & \multicolumn{1}{c}{{[\si{\milli \meter}]}} & \multicolumn{1}{c}{{[\si{\kelvin}]}} \\
		\midrule
		{Mean} 				& 267  & 620	& 128 	& 119	& \num{16402} 	& 42  & 5.4 & 4.5 & 3.8 & 1.0 & 741 \\
		{Std}  				& 93  & 343		& 44 	& 57	& \num{7070} 	& 21  & 3.9 & 2.1 & 1.9  & 0.1 & 177 \\
		\bottomrule
	\end{tabular}
	\caption{Mean value, standard deviation and percentiles of the test data}
	\label{tab:properties_test}
\end{table}

\end{document}